\pdfoutput=1

\documentclass[11pt]{article}

\usepackage{EMNLP2022}

\usepackage{times}
\usepackage{latexsym}

\usepackage[T1]{fontenc}

\usepackage[utf8]{inputenc}

\usepackage{inconsolata}

\usepackage{microtype}
\usepackage{pifont}
\usepackage{latexsym}

\usepackage{balance}
\usepackage{helvet}  
\usepackage{courier}  
\usepackage{url}  
\usepackage{graphicx}  
\usepackage{amssymb}
\usepackage{amsmath}
\usepackage{algorithm}
\usepackage{algorithmic}
\usepackage{amsthm}
\usepackage{multirow}
\usepackage{pgffor}
\usepackage{graphicx}
\usepackage{epstopdf}
\usepackage{tikz}
\usepackage{epstopdf}
\usepackage{mathtools}
\usepackage{enumitem}
\usepackage{subfigure}
\usepackage{tabulary}
\usepackage{color}
\usepackage{url}
\usepackage{flushend}
\usepackage{stfloats}
\usepackage{tcolorbox}
\usepackage[utf8]{inputenc}
\usepackage[english]{babel}
\usepackage{tabularx}
\usepackage{CJKutf8}
\usepackage{microtype}
\usepackage{textcomp}
\usepackage{booktabs}
\usepackage{colortbl}
\usepackage{soul}
\usepackage{arydshln}
\usepackage{natbib}
\usepackage{appendix}

\title{Uni-Parser: Unified Semantic Parser for Question Answering on Knowledge Base and Database}

\author{
  \textbf{Ye Liu}$^1$, \textbf{Semih Yavuz}$^1$, \textbf{Rui Meng}$^1$, \textbf{Dragomir Radev}$^2$, \textbf{Caiming Xiong}$^1$, \textbf{Yingbo Zhou}$^1$ \\
$^1$ Salesforce Research, $^2$ Yale University\\
  { \texttt{\{yeliu, syavuz, ruimeng, yingbo.zhou, cxiong\}@salesforce.com},} \\ {\texttt{dragomir.radev@yale.edu}}
}

\newcount\DraftStatus  
\definecolor{darkgreen}{rgb}{0,0.5,0} 
\definecolor{purple}{rgb}{1,0,1} 
\definecolor{todocolor}{rgb}{0.9,0.1,0.1} 
\definecolor{fixcolor}{rgb}{0.1,0.7,0.3} 
\definecolor{wycolor}{rgb}{0.9,0.1,0.1} 
\definecolor{hycolor}{rgb}{0.7,0.7,0.3} 

\newcommand{\nbc}[3]{\ifnum\DraftStatus=1
	{\colorbox{#3}{\bfseries\sffamily\scriptsize\textcolor{white}{#1}}}
	{\textcolor{#3}{\sf\small$\blacktriangleright$\emph{#2}$\blacktriangleleft$}}
\fi}

\newcommand{\draftnote}[2]{\ifnum\DraftStatus=1
	\marginpar{
		\tiny\raggedright
		\hbadness=10000
		\def\baselinestretch{0.8}
		\textcolor{#1}{\textsf{\hspace{0pt}#2}}}
\fi}

\begin{document}
\maketitle
\begin{abstract}
Parsing natural language questions into executable logical forms is a useful and interpretable way to perform question answering on structured data such as knowledge bases (KB) or databases (DB). 
However, existing approaches on semantic parsing cannot adapt to both modalities, as they suffer from the exponential growth of the logical form candidates and can hardly generalize to unseen data.
In this work, we propose \textbf{Uni-Parser}, a unified semantic parser for question answering (QA) on both KB and DB.
We introduce the \textit{primitive} (\textit{relation} and \textit{entity} in KB, and \textit{table name}, \textit{column name} and \textit{cell value} in DB) as an essential element in our framework.  
The number of primitives grows linearly with the number of retrieved relations in KB and DB, preventing us from dealing with exponential logic form candidates. 
We leverage the generator to predict final logical forms by altering and composing top-ranked primitives with different operations (\textit{e.g. select, where, count}). 
With sufficiently pruned search space by a contrastive primitive ranker, the generator is empowered to capture the composition of primitives enhancing its generalization ability.
We achieve competitive results on multiple KB and DB QA benchmarks more efficiently, especially in the compositional and zero-shot settings. 


\end{abstract}

\section{Introduction}
With the recent advances in deep neural networks, question answering (QA) systems enable users to interact with massive data using queries in natural language. 
However, it remains challenging to assess structured data, such as knowledge bases and databases.
Semantic parsing is a core step of question answering for structured data. The goal is to convert a natural language question to an executable logical form~\cite{berant2013semantic,yih2015semantic}, e.g., SQL for databases and S-expression for knowledge bases. 

To improve the accuracy and faithfulness in execution of semantic parsing, recent KBQA studies propose to generate logical form candidates by enumerating and selecting the best logical form by ranking~\cite{berant2014semantic, yih2015semantic,sun2020sparqa,ye2021rng}. 
However, the number of logical form candidates may grow exponentially with the increase of reasoning depth for complex questions. 
Thus this approach can suffer from poor runtime performance due to the time-consuming logical form enumeration~\cite{gu2021beyond} and inefficient candidate ranking~\cite{ye2021rng}. 
For example, given an entity in a KB, we collect its logical form candidates by enumerating paths up to two hops. 
If the first hop with respect to the given entity contains N relations and the second hop contains M relations, the enumeration would result in $N\times M$ logical forms. As KBs typically contain massive structured knowledge, an entity can have hundreds of linked relations~\cite{bollacker2008freebase}. 
Moreover, for complex questions, which require combining logic or aggregation operations, such as COUNT, ARGMIN, or ARGMAX, into the logical form, the situation could be even worse.
Therefore, traditional enumeration methods may fail in cases requiring complicated reasoning (e.g., involving large amounts of entities and long reasoning chains). 
The situation becomes more severe and prohibitive if one wants to apply the enumeration method to other structured data with dense connections between entities such as in databases.
Designing a unified semantic parsing method for various modalities of structured data have significant theoretical and practical value, yet it is still an understudied topic.




To avoid the problem of exponential growth in the logical form enumeration, we consider logical forms composed of two types of elements -- primitives and operations. Primitives are defined by the schema of the structured data source and operations are a set of grammars associated with primitives. For example, in the context of knowledge bases, primitives are defined as \textit{relations} and \textit{entities} in the knowledge graph. Whereas in databases, primitives are presented as \textit{tables}, \textit{columns}, and \textit{cells}. Through this formulation, the number of candidates can be greatly reduced, going down from $N\times M$ to $N+M$.

In this study, we present \textbf{Uni-Parser}, a unified semantic parser for question answering on both knowledge bases (KBs) and databases (DBs).  
Our model follows the framework of Enumeration-Ranker-Generator proposed in RnG-KBQA~\cite{ye2021rng}. 
We first enumerate possible question-relevant primitives of a given KB or DB. Then a cross-encoder ranker is utilized to select the best candidates with contrastive learning~\cite{chang2020pre}, and it is further enhanced by a special hard negative sampling strategy.
After getting the top-k ranked primitives for each hop, we filter out the high-order primitives that cannot be reached from the KB or do not exist in the DB through selected low-order primitives. 
Next, we introduce a generator that consumes both the question and filtered top-k primitives with predicted operations to compose the final logical form. 
Starting from primitives rather than logical forms, our generator needs to understand the semantic meaning of each primitive to compose them into the logical form. 

Our contributions can be summarized as follows:  \\
\noindent$\bullet$ We propose a unified semantic parser working for both KB and DB question answering.\\
\noindent$\bullet$ We enumerate primitives rather than logical forms, which greatly reduces the search space and makes candidate generation and ranking more efficient and scalable. \\
\noindent$\bullet$ The composition of logical forms from primitives and operations is postponed to the generation phase. Thus, the generator is required to learn the compositional relations among primitives. This leads to a more generalized model that can work on complex logical forms and generalize to questions involving unseen schema. \\
\noindent$\bullet$ Extensive empirical results on four KB and DB QA datasets demonstrate the effectiveness, flexibility and scalability of our unified framework.

\section{Problem Formulation}
Given a structured data source $D$ and a question $X$ in natural language, a semantic parser model is tasked to generate the corresponding logical form $Y$. Specifically, we illustrate the details depending on the type of $D$ as follows:
(1) \textbf{Knowledge Base:} Data is stored in the form of subject-relation-object $(s, r, o)$, where $s$ is an entity, $r$ is a relation and $o$ is an entity or a literal (e.g., integer values, data, etc.). We use S-expressions~\cite{gu2021beyond} to represent logical forms for KB. S-expression is used to query a KB with the entity type, and the operations on the KB are treated as operations on a set of entities. This formulation greatly reduces the number of operations in traditional lambda DCS~\cite{liang2013lambda}. 
(2) \textbf{Database:} A DB QA dataset typically consists of multiple tables $T = \{t_1, \cdots, t_N\}$. Each table $T_i$ contains $M$ columns: $C = \{c_1, \cdots, c_M\}$. Each column includes multiple cell values $V = \{v_1, \cdots, v_L\}$, where $L$ is the number of rows. For tabular data, SQL is used to represent logical forms~\cite{yu2018spider}.

Logical forms can be decomposed into the primitives and operations, both of which are defined by the schema of the structured data. We define \textbf{primitives} as the atomic elements that are entities themselves or can be used to navigate to entities.
\textbf{Operations} are a set of grammars to associate primitives. Thus, a logical form can be decomposed to primitives and operations. We list specific primitives and operations considered in this study in Table~\ref{operation_tb}.
\begin{table}[]
\centering
\resizebox{0.48\textwidth}{!}{
\begin{tabular}{ll}
\toprule
\textbf{KB} & \textbf{DB} \\ \toprule
\textbf{Primitive on S-expression}      & \textbf{Primitive on SQL}    \\ \midrule

Relation, Entity    &        Table Name, Column Name, Cell Value \\ \midrule
\midrule
\textbf{Operation on S-expression}                                           & \textbf{Operation on SQL}                                                      \\ \midrule

\multicolumn{2}{l}{\textbf{\textit{Logical Operation}}} \\ \midrule

AND,  JOIN, R                                              & SELECT, WHERE, ORDER/GROUP BY   \\ \midrule

\multicolumn{2}{l}{\textbf{\textit{Aggregation Operation}}}  \\\midrule 
COUNT, ARGMAX, ARGMIN                                      & AVG, COUNT, MAX, MIN, SUM                                               \\ \midrule

\multicolumn{2}{l}{\textbf{\textit{Conditional Operation}}} \\ \midrule

\textless{}, \textless{}=, \textgreater{}, \textgreater{}= & between, =, \textgreater{}, \textless{}, \textgreater{}=, \textless{}=, \\
                                                           & !=, in, like, is, exists, not in,                                       \\
                                                           & not like, not between, is not                                           \\ \midrule

\multicolumn{2}{l}{\textbf{\textit{Conjunction Operation}}} \\ \midrule
~& and, or, except, intersect, union \\ 

\bottomrule                                                          
\end{tabular}
}
\caption{Primitives and Operations in KB and DB logical form.}
\label{operation_tb}
\end{table}
 
\section{Methodology}
\begin{figure}[t]
\centering
\includegraphics[width=1\linewidth]{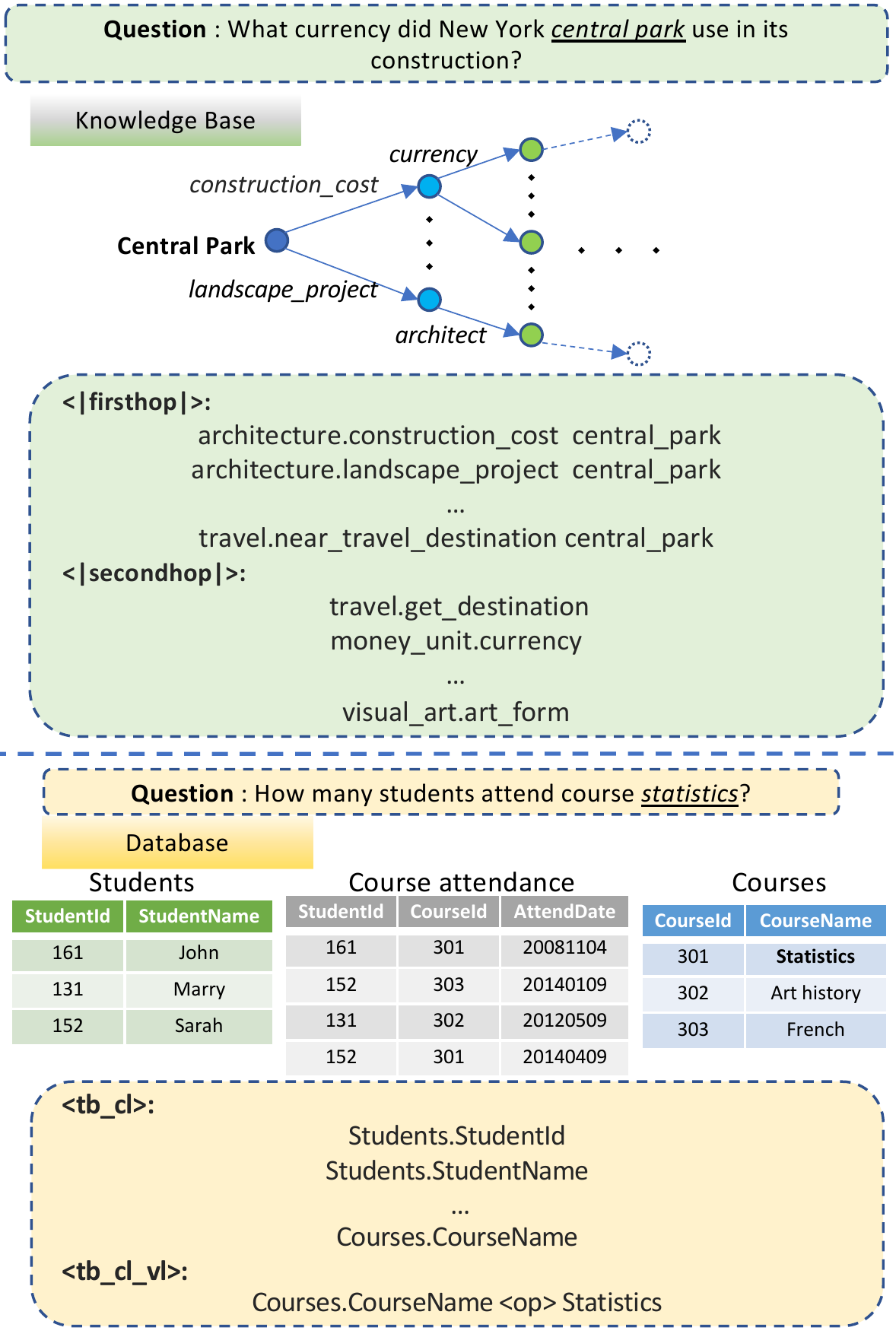}
\caption{Given the question and its knowledge source, the primitive enumeration process produces different categories of primitives for KB and DB.}
\label{figure:primitive}
\end{figure}
Our model first obtains the relevant primitives from the given structured data and question, and sets them into different categories (Sec \ref{sec:enum}). Then a ranker filters out the irrelevant primitives (Sec \ref{sec:rank}) and provides the top-ranked primitives to the generator to produce the final logical form (Sec~\ref{sec:generator}). We'll first explain how to extract the primitive candidates from KB or DB based on the question. 

\subsection{Primitive Enumeration} \label{sec:enum}
Rather than enumerating all possible logical forms as in RnG-KBQA~\cite{ye2021rng}, here we only enumerate primitives that are relevant to the question.

In cases of knowledge bases, we start by detecting the entity mentioned in the question with the help of an out-of-the-box NER system and then run fuzzy matching \cite{lin2020bridging} with the entity names in the knowledge base to identify relevant entities. This technique is also used in~\cite{gu2021beyond,chen2021retrack}. To alleviate the issue of entity disambiguation, we follow~\cite{ye2021rng} to use a ranker model to select entity candidates based on the similarity between the question and the one-hop in/out relations of the entity. Nevertheless, since most questions contain two-hop reasoning in KBs, we also extract two-hop paths associated with question entities as a related sub-graph. We define \texttt{<|firsthop|>} category primitives as the entities with first-hop relation and likewise for \texttt{<|secondhop|>} category primitives (examples are shown at the top of Figure \ref{figure:primitive}). 

As for databases, we consider two formats of primitives. 
The first category is \texttt{<|tb\_cl|>}, denoting the format table$\_$name.column$\_$name. 
The second one is \texttt{<|tb\_cl\_vl|>}, representing the format table$\_$name.column$\_$name <op> cell$\_$value. 
The $\texttt{<op>}$ represents a conditional operation as shown in Table \ref{operation_tb}. 
To enumerate the first category of primitives, we can simply use all table names together with their column names. 
However, for the second category, including cell values in enumeration will lead to a vast amount of candidates. 
For instance, on the Spider dataset~\cite{yu2018spider}, if we treat every cell value as a candidate, we can get up to 263K candidates for one question. To address this issue, following~\cite{lin2020bridging}, we perform a fuzzy string match between question $X$ and the cell value $V$ under each column name $C$, and pair the matched value with its corresponding column name. One shortcoming of using string match is, that we can only obtain coverage of 15$\%$ for the Spider dataset. This is because many cell values are of numeric type, where string match fails to detect. For example, for the question "How many heads of departments are older than 56?", there is no cell value that matches 56. Therefore, if a question contains numbers, we pair them with all column names in the table.    


\begin{figure}[t]
\centering
\includegraphics[width=1\linewidth]{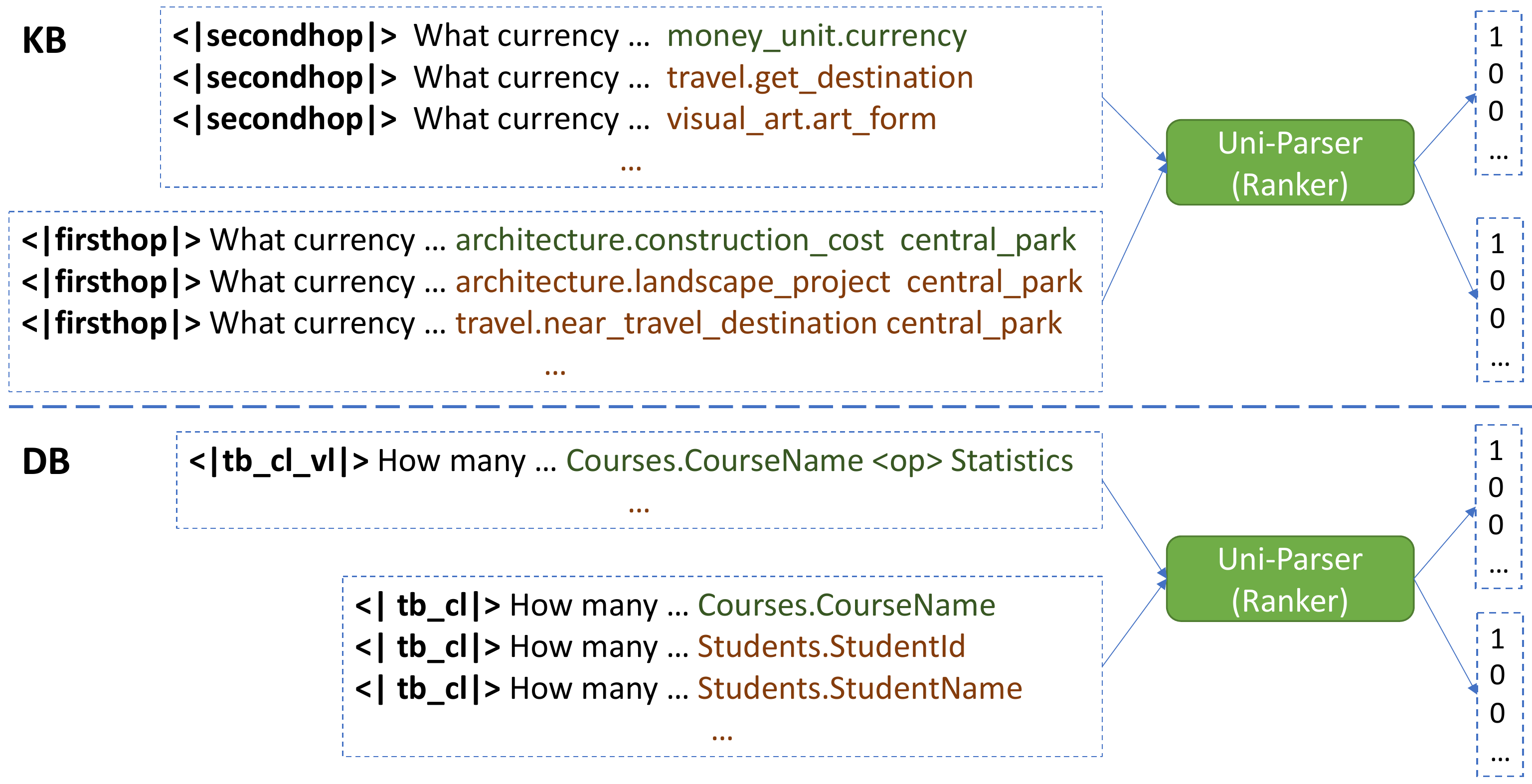}
\caption{An illustration of Uni-Parser ranker.}
\label{figure:ranker}
\end{figure}

\subsection{Primitive Ranker} \label{sec:rank}
Our ranker model learns to filter out irrelevant primitives by measuring the similarity between questions and primitive candidates. 
We utilize the cross-encoder architecture~\cite{chang2020pre} for ranker, which has shown to be more performant than bi-encoder architectures~\cite{thakur2020augmented,lei2022loopitr}. 
As shown in Figure \ref{figure:ranker}, we use a special category token as a prompt in the input to differentiate primitives in the input. 
Note that all primitives from different categories share the same ranker. 
Specifically, given a question $X$ and a primitive $p$ with a category token $p_c$, we use a BERT-based encoder that takes as input the concatenation of their vector representations, and outputs a logit representing the similarity between the primitive and the question:
\begin{align}
    s(X, p, p_{c})= \operatorname{FFN}(\psi_{\theta}(p_c \oplus X \oplus p))
\end{align}
where $\oplus$ denotes a concatenation operation. $\psi_{\theta}$ denotes the [CLS] representation of the concatenated input after BERT embedding; $\operatorname{FNN}$ is a projection layer reducing the representation to a scalar similarity score. $p_c$ is the special token to distinguish the category of the primitive (in KB, $p_c \in \{$\texttt{<|firsthop|>, <|secondhop|>}$\}$ and in DB, $p_c \in \{$\texttt{<|tb$\_$cl|>, <|tb$\_$cl$\_$vl|>}$\}$).

The ranker is optimized to minimize the contrastive loss:
\begin{align}
    \mathcal{L}_{\theta}\left(X, ~p^{+}, \mathcal{P}^{-}, p_{c}\right)=-\frac{e^{s\left(X, ~p^{+}, ~p_{c}\right)}}{\sum_{p \in\left\{p^{+}\right\} \cup \mathcal{P}^{-}} e^{s\left(X, ~p, ~p_{c}\right)}}
\end{align}
where $p^{+}$ is the positive primitive extracted from the gold logical form and $\mathcal{P}^{-}$ is the set of negative primitives from the same category $p_{c}$.

\subsubsection{Negative Sampling} 
Since a large number of negative primitive candidates can be paired with a positive example, it is necessary to apply negative sampling.
A straightforward way for this is random sampling, however it may suffer from the domination of uninformative negatives~\cite{xiong2020approximate}. To this end, we design a strategy to sample hard negative candidates for training the ranker~\cite{liu2021dense}. In cases of KBs, the number of second hop relations can grow exponentially compared to the first hop. Thus the hard negative candidates of the second hop can only be sampled from the primitives connected to the ground truth first hop. In cases of DBs, for \texttt{<|tb$\_$cl|>} category primitives, we treat those having the same table name with ground truth but different column names as the hard negatives. And for the \texttt{<|tb$\_$cl$\_$vl|>} category, we treat candidates with the same table and column name with ground truth, but having a different cell value as the hard negatives. 
Moreover, the bootstrap negative sampling strategy is leveraged; that is, the model is trained recursively using the false positive candidates generated from the last training epoch.

\subsubsection{Primitive Candidates Filtering}
In KBs, the top ranked first hop primitives and second hop primitives can be formed into two-hop paths by combining one first hop primitive with each of the second hop. 
However, the resulting paths may not exist in the KB. To provide valid primitive candidates to the generator, we filter out the second hop primitives that cannot be reached from any of the first hop primitives.

\begin{figure}[t]
\centering
\includegraphics[width=1\linewidth]{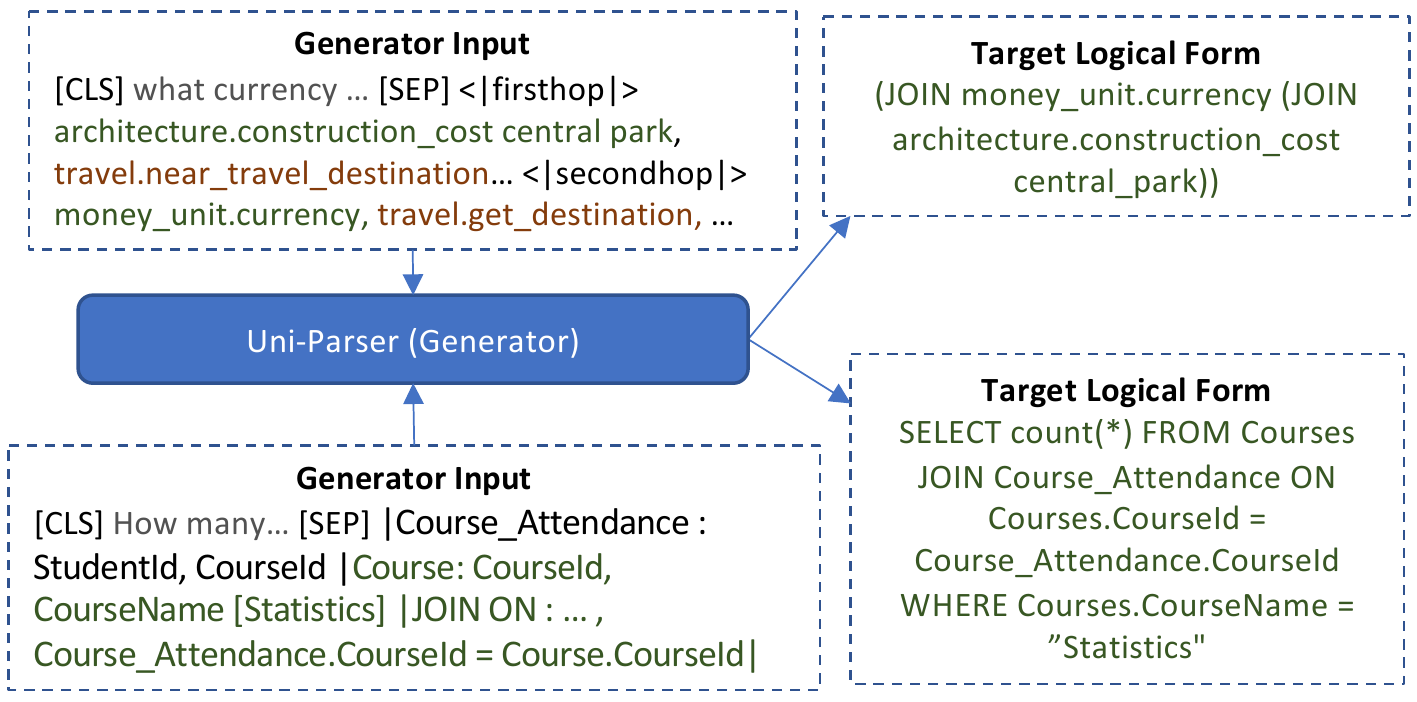}
\caption{An illustration of Uni-Parser generator.}
\label{figure:generator}
\end{figure}

\subsection{Logical Form Generation by Composing Primitives}  \label{sec:generator}
In the final stage, we need to employ a generator model to predict the target logical form by composing the top primitive candidates provided by the ranker.
We use a T5 model~\cite{raffel2020exploring} as the basis of our logical form generator, as it demonstrates strong performance on various text generation tasks. We construct the inputs by concatenating the question and the top-k primitive candidates. As shown in Figure~\ref{figure:generator}, the input for KBs is formatted as:
\texttt{[X; <|second\_hop|> primitives ; <|first\_hop|> primitives]}. As for DBs, its input is formatted as: \texttt{[X; |table\_name$_{1}$| column\_name$_{1}$, column\_name$_{2}$ <op> value $\cdots$ |table\_name|$_{2}$ $\cdots$}.
We train the model by teacher forcing -- the target logical form is generated token by token and the model is optimized with the loss of cross-entropy. At inference time, we use beam-search to decode top-k target logical forms in an autoregressive manner. 

During training, it is often the case that the ranker performs well and predicts the gold primitive at the top place, but this positional information can be misused by the generator. If the generator is biased by the ranking of primitives, it may generalize poorly to unseen data. In order to encourage the generator to focus on the semantic meaning of each primitive rather than their positions, we shuffle the order of primitives in the input during training. According to our experiments, we find that the generator can benefit from this shuffle augmentation and perform robustly against the noise in the ranked primitives. 


Our generator learns to generate logical forms by understanding the meaning of its elements -- primitives and operations -- and composing them. Compared to RnG-KBQA~\cite{ye2021rng}, whose generator predicts logical forms based on a list of logical form candidates, the compositionality on the basis of primitives and operations can make our model likely to generalize better on unseen structured data.


\section{Experiments}
\subsection{Dataset and Evaluation}
\textbf{KBQA}
(1) GRAILQA \cite{gu2021beyond} contains 64,331 questions and carefully splits the data to evaluate three levels of generalization in the task of KBQA, including i.i.d. setting, compositional generalization to unseen composition of KB schema, and zero-shot generalization to unseen KB schema. The fraction of each setting in the test set is 25$\%$, 25$\%$, and 50$\%$, respectively. 
(2) WebQSP \cite{yih2016value} is a dataset that evaluates KBQA approaches in i.i.d. setting. It contains 4,937 questions and requires reasoning chains with up to 2 hops. Similar to \citet{ye2021rng}, we randomly sample 200 examples from the training set for validation. \\
\textbf{DBQA}
(1) Spider \cite{yu2018spider} is a multi-table text-to-SQL dataset, which contains 10,181 questions and 5,693 complex SQL queries on 200 databases. There is no overlap between train/dev/test databases. (2) WikiSQL \cite{zhong2017seq2sql} is a single-table text-to-SQL dataset, which contains 80,654 questions and SQL queries distributed across 24,241 tables from Wikipedia. 49.6$\%$ of its dev tables and 45.1$\%$ of its test tables are not in the training set. Therefore, both datasets require models to generalize to the unseen schema composition (compositional generalization) and unseen schema (zero-shot generalization).

\textbf{Evaluation Metrics.}
We use their official evaluation script for each dataset with two metrics to measure logical form of program exact match accuracy (EM) and answer accuracy (F1).

\subsection{Results on KBQA}
We first test our approach on KBQA with the GrailQA and WebQSP datasets. 

\begin{table*}[t]
\centering
\resizebox{0.95\textwidth}{!}{
\begin{tabular}{lcccccccc}
\toprule
                  & \multicolumn{2}{c}{Overall} & \multicolumn{2}{c}{I.I.D} & \multicolumn{2}{c}{Compositional} & \multicolumn{2}{c}{Zero-Shot} \\ \cline{2-9} 
                  & EM           & F1           & EM          & F1          & EM              & F1              & EM            & F1            \\ \hline
Bert Ranking(Test)  \cite{gu2021beyond}    &    50.6          &      58.0        &      59.9       &    67.0         &   45.5              &      53.9           &     48.6          &     55.7          \\
ReTrack (Test) \cite{chen2021retrack}         &    58.1          &      65.3        &     84.4        &      87.5       &      61.5           &      70.9           &   44.6            & 52.5   \\
UnifiedSKG (Test) \cite{xie2022unifiedskg}   &     62.4       &   -    &       -     &   -    &      -    &    -   &      -     &   -        \\
RNG-KBQA (Test)  \cite{ye2021rng}        &     68.8         &     74.4         &      \textbf{86.2}       &      \textbf{89.0}       &       63.8          &          \textbf{71.2}       &      63.0         &    69.2           \\   \hdashline
UnifiedSKG (Dev) \cite{xie2022unifiedskg}   &     60.0       &   -    &       -     &   -    &      -    &    -   &      -     &   -        \\
RNG-KBQA (Dev)  &     70.7       &    75.7    &    \textbf{86.4}  &  \textbf{88.6}     &       61.7    &   68.3      &  67.4       &    73.1       \\  \hline
Uni-Parser (Dev)    &       \textbf{70.8}     &   \textbf{76.5}     &     85.7       &  88.3    &     \textbf{62.8}      &  \textbf{71.4}      &    \textbf{67.7}     &    \textbf{73.4}       \\
Uni-Parser (Test)    &       \textbf{69.5}     &   \textbf{74.6}     &     85.5       &  88.5    &     \textbf{65.1}      &  71.1      &    \textbf{64.0}     &    \textbf{69.8}       \\
\bottomrule        
\end{tabular}
}
\caption{Exact match (EM) and F1 scores on test/dev split of the GRAILQA. The numbers of the baselines are taken from leaderboard and their research works. The reported models are based on BERT-base model for ranker and T5-base for generator. Best results among dev are bolded and the results of test better than dev are underlined.}
\label{grailqa}
\end{table*}


\begin{table}[]
\centering
\resizebox{0.45\textwidth}{!}{
\begin{tabular}{lcc}
\toprule
         & \textbf{EM}  &  \textbf{F1}   \\ \hline
Topic Units \cite{lan2019knowledge}   & -  & 67.9      \\
STAGG \cite{yih2015semantic} & 63.9   & 71.7     \\
QGG  \cite{lan2020query}    & -   & 74.0    \\
CBR  \cite{das2021case}     & 70.0  & 72.8  \\
ReTrack \cite{chen2021retrack}  & -   & 71.0   \\
RNG-KBQA \cite{ye2021rng}  & 71.1 & 75.6 \\
ArcaneQA \cite{gu2022arcaneqa}     & -   &  75.3    \\ \hline
Uni-Parser     &    \textbf{71.4}  &  \textbf{75.8}    \\
\bottomrule
\end{tabular}
}
\caption{Exact match (EM) and F1 scores on the test split of WebQSP. The reported models are based on BERT-base model for ranker and T5-base for generator.}
\label{webqsp}
\end{table}

\subsubsection{Implementation Details}
For GrailQA and WebQSP, we use the entity linking results provided by \cite{ye2021rng}. After identifying a set of entities, we extract the primitives within 2 hops from the question entities. We initiate the primitive ranker using BERT-base-uncased. For each primitive category, 96 negative candidates are sampled. We trained the ranker for 3 epochs using a learning rate of 1e-5 and a batch size of 8. Bootstrap sampling is applied after every epoch. It is also noteworthy that we perform teacher-forcing when training the ranker, i.e., we use ground truth entity linking for enumerating training candidates. 
We base our generation model on T5-base \cite{raffel2020exploring}. We use top-10 primitives from each category returned by the ranker and finetune the T5 generation model for 10 epochs using a learning rate of 3e-5 and a batch size of 8. 
A vanilla T5 generation model is used without syntactic constraints, which does not guarantee the syntactic correctness nor executability of the produced logical forms. Therefore, we use an execution-augmented inference procedure, which is commonly used in previous semantic parsing related work \cite{devlin2017robustfill,ye2020sketch}. We first decode top-k logical forms using beam search and then execute each logical form until finding one that yields a valid (non-empty) answer. In case none of the top-k logical forms is valid, the top-ranked primitives obtained using the ranker is returned. The rule-based method is used to formulate the final logical form, which is guaranteed to be executable. This inference schema can ensure finding one valid logical form for each problem. 
\subsubsection{Overall Evaluation}
Table \ref{grailqa} and \ref{webqsp} summarize the results on GrailQA and WebQSP, respectively. Our approach achieves the highest overall performance among all approaches. Compared with the methods enumerating logical forms like Bert Ranking and RnG-KBQA, our approach achieves better performance in compositional and zero-shot settings. Especially, we get 3.1$\%$ improvement over baselines on F1 on dev set and 1.3$\%$ improvement over baselines on EM on test set in the compositional setting. This matches our expectation that our generator learns the composition of the primitives. RnG-KBQA enumerates logical forms rather than primitives. Therefore its generation module acts more like an auxiliary input to complement the enumerated logical forms. However, the rationale behind our generation module is to compose logical forms with basic semantic units(primitives). Thus, with the sense of primitive composition, our model is more capable of dealing with unseen composition than RnG-KBQA. For i.i.d. setting, our approach underperforms RnG-KBQA by 0.7$\%$ on EM and 0.5$\%$ on F1. We speculate that our model needs to understand whether a question implies a one or two hops reasoning on the KB, which is a difficult task. But the generator of RnG-KBQA already sees logical form candidates in the input, it does not need to deal with this problem.

\subsubsection{Efficiency Analysis}
We compare the running time of the Uni-Parser and ranking-based model. To make a fair comparison, we time the process from the enumeration step to the logical form generation, using 1,000 randomly sampled questions on GrailQA datasets. We also report the average running time per question on an A100 GPU. Our model uses 19.4s, which is considerably faster than BERT+Ranking (76.3s) and RnG-KBQA (53.5s). 
Unlike logical form-based models, our model doesn't need to enumerate a large amount of logical forms. Instead, only a small number of relevant primitives are considered, which leads to faster tokenization and efficient ranking.



\begin{table}[]
\centering
\resizebox{0.48\textwidth}{!}{
\begin{tabular}{lllll}
\toprule
              & \textbf{EM}   & \textbf{F1}   \\ \hline
Global-GNN \cite{bogin2019global}          & 52.7 & -  \\
EditSQL    \cite{zhang2019editing}      & 57.6 & -     \\
T5-Base  \cite{scholak2021picard}    & 57.2    &  57.9  \\
UnifiedSKG(T5-Base) \cite{xie2022unifiedskg}   & 58.1   &  - \\ \hline
Uni-Parser(T5-Base)  & \textbf{61.2} & \textbf{62.8}\\ \toprule
RAT-SQL \cite{wang2019rat} & 69.7 & - \\
BRIDGE(Large) \cite{lin2020bridging}  & 70.0 & 68.0  \\
T5-3B*  \cite{scholak2021picard}    & 71.5    &  \textbf{74.4}  \\
UnifiedSKG(T5-3B) \cite{xie2022unifiedskg} & \textbf{71.7}   &  - \\\hline
Uni-Parser(T5-3B) & 71.2 & 71.8  \\
\bottomrule  
\end{tabular}
}
\caption[{https://github.com/ServiceNow/picard/blob/main/configs/train.json}]{Exact match (EM) and F1 scores on the dev split of Spider. The upper block shows the comparison of the small pre-trained models, and the lower block shows the comparison of the large pre-trained models. Our model trains T5-3B model 100 epochs while T5-3B * trains 3072 epochs \footnotemark.}
\label{spider}
\end{table}
\footnotetext{\url{https://github.com/ServiceNow/picard/blob/main/configs/train.json}}
\begin{table}[]
\centering
\resizebox{0.48\textwidth}{!}{
\begin{tabular}{lccc}
\toprule
               & \textbf{EM}   & \textbf{F1}   \\ \hline
SQLova   \cite{hwang2019comprehensive}      & 80.7 & 86.2  \\
X-SQL   \cite{he2019x}         & 83.3 & 88.7      \\
IE-SQL  \cite{ma2020mention}  & 84.6 & 88.8  \\
NL2SQL  \cite{guo2019content}   & 83.7    & 89.2   \\
HydraNet  \cite{lyu2020hybrid}  & 83.8    & 89.2    \\
BRIDGE(Large) \cite{lin2020bridging}        & 85.7    & 91.1    \\ 
TAPEX \cite{liu2021tapex} & - & 89.5 \\ \hline
Uni-Parser(T5-Base)  & 85.8 & 91.3 \\
Uni-Parser(T5-Large)  & \textbf{86.9} & \textbf{92.1} \\
\bottomrule
\end{tabular}
}
\caption{Exact match (EM) and F1 scores on the test split of WikiSQL.}
\label{wikisql}
\end{table}

\subsection{Results on DBQA}
We also evaluate our approach to the DBQA task with Spider and WikiSQL datasets. 
\subsubsection{Implementation Details}
To construct the <|tb$\_$cl$\_$vl|> category primitives mentioned in Section \ref{sec:enum}, we find the relevant cell values related to the question. Given a question and DB, we compute the string matching between the arbitrary length of phrase in question and the list of cell values under each column of all tables. We followed \cite{lin2020bridging} to use a fuzzy matching algorithm to match a question to a possible cell value mentioned in the DB. We also detect the number value in the question and form all column names with the value as the primitives. 
We find that the column name in the WikiSQL dataset is vague, like ``No.'', ``Pick \#'', and ``Rank'', so we use the cell value to supplement the meaning of the column name. We use the matching cell value to locate the row and match the column name with the cell value in the same row. 

We initiate the primitive ranker using BERT-base-uncased. We sample 48 negative candidates for each primitive category. We trained the ranker for 10 epochs using a learning rate of 1e-5 and a batch size of 8. Bootstrap hard negative sampling is conducted after every two epochs. We also use ground truth entity linking for enumerating training candidates. For the generator, we trained it using T5-base and 3B on Spider datasets. We use top-15 <|tb$\_$cl|> category primitives and top-5 <|tb$\_$cl$\_$vl|> category primitives returned by the ranker and finetune the T5-base model for 200 epochs using a learning rate of 5e-5 and a batch size of 64. For the T5-3B model, we run it on 16 A100 GPUs with 100 epochs using a batch size of 1024. And on the WikiSQL dataset, we use T5-base and T5-large, and use top-5 <|tb$\_$cl|> category primitives and top-3 <|tb$\_$cl$\_$vl|> category primitives as the input of the generator. We finetune the T5-base/large model for 20 epochs using a learning rate of 3e-5 and a batch size of 16.  

\subsubsection{Overall Evaluation}
Table \ref{spider} and \ref{wikisql} summarize the results on Spider and WikiSQL respectively.
On the challenging Spider dataset, our model achieves competitive performance among all baseline models. Compared with generation models that use whole DB table schema as input like BRIDGE and UnifiedSKG on T5-base models, our model achieves 3$\%$ improvement, suggesting the advantage of our method. During primitive enumeration and ranking, we filter out irrelevant candidates based on the DB table schema. Compared with the other T5-3B models, our model achieves comparable performance with fewer training epochs where T5-3B* trained ~3K epochs, while we train only 100 epochs. For WikiSQL, we compare the Text2SQL methods and answer generation method (TAPEX) in Table \ref{wikisql}, and Uni-Parser outperforms all the baselines.  
\begin{table}[]
\centering
\resizebox{0.49\textwidth}{!}{
\begin{tabular}{lccc}
\toprule
      & \textbf{w/o CG} & \textbf{w/o HN} & \textbf{w/ HN} \\ \hline
\textbf{KBQA: WebQSP} & & & \\ 
Top-1  & 54.5        & 56.2             & \textbf{56.2}     \\
Top-10 & 67.8        & 68.8             & \textbf{69.0}     \\
Top-20 & 70.4        & 71.4             & \textbf{71.8}     \\ \hline
\textbf{TableQA: WikiSQL} & & & \\ 
Top-1  & 80.5        & 81.9             & \textbf{84.0}     \\
Top-3 & 86.5        & 87.9             & \textbf{91.6}     \\
Top-5 & 88.4        & 90.0             & \textbf{94.0}     \\
\bottomrule
\end{tabular}
}
\caption{The Recall of the top-K ranked primitive on KB and Table. CG means category, HN means Hard Negative.}
\label{ranker}
\end{table}
\begin{figure*}[t]
\centering
\includegraphics[width=1\linewidth]{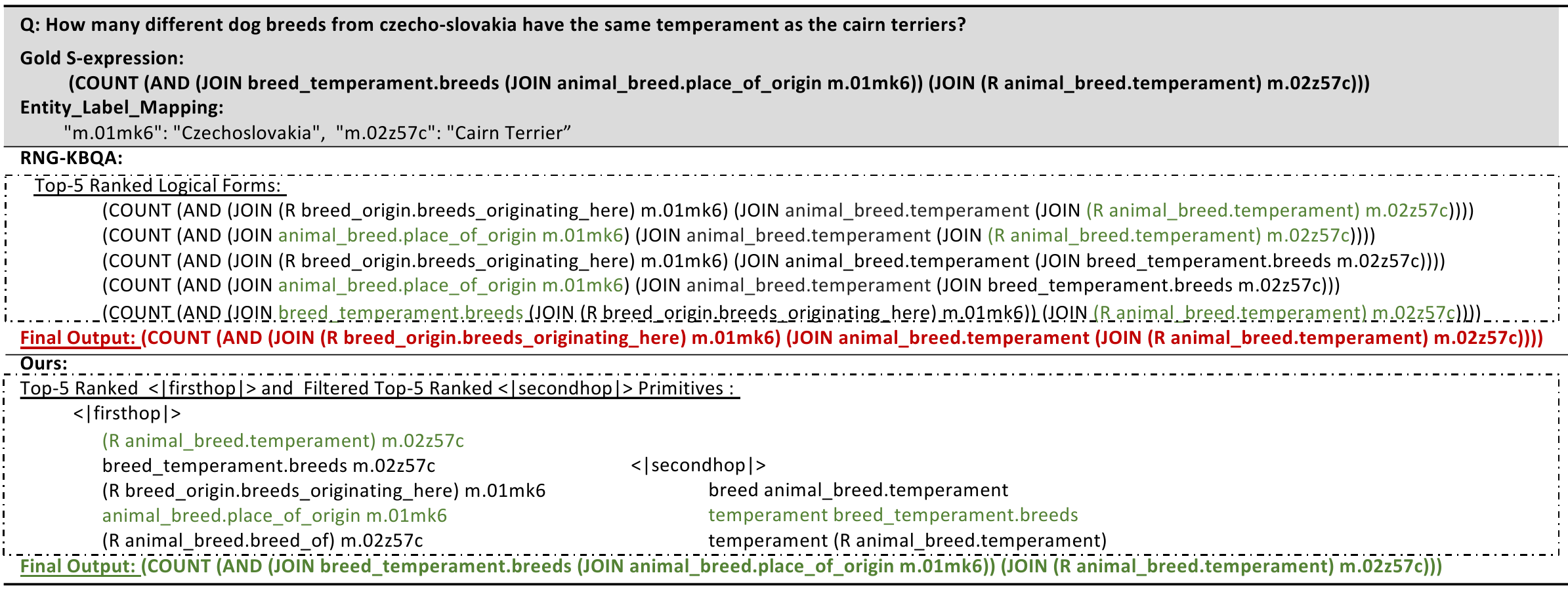}
\caption{Ranker output (shown in dotted boxes) and Generator output (shown in Final output) comparison between our primitive-based method Uni-Parser and logical form-based method RnG-KBQA on GrailQA dev set. Our model generates the correct output while the logical form-based model produces a wrong output that is the same as the top-1 ranked output.}
\label{figure:case_study}
\end{figure*}

\subsection{Analysis}
~~~~\textbf{Ablation Study} We perform an ablation study on the effects of hard negative strategies in ranking on WebQSP and WikiSQL, and the result is shown in Table \ref{ranker}. No CG means that we do not differentiate primitives by their categories. The intention of this design is to help the ranker to distinguish whether a question has two or one hop reasoning. No CG shows lower performance than the setting using categories in the input. 
By comparing the settings of with and without hard negative (rightmost two columns), we can see that the proposed hard negative sampling can help the ranker to better determine the positive primitive from the negative ones. Moreover, the accuracy of entity linking is WebQSP is 72.5, which means the upper bound of the ranking stage. Therefore, our model achieves 69.0 among Top-10 primitive is close to the oracle performance. 

\begin{table}[]
\centering
\resizebox{0.49\textwidth}{!}{
\begin{tabular}{ll|c|cc}
\toprule
\multicolumn{2}{l|}{\textbf{Dataset}}                      & \textbf{LF} & \multicolumn{2}{c}{\textbf{Primitive}}                                             \\ \hline
\multicolumn{1}{c}{\multirow{3}{*}{\textbf{KB}}} &         &              & <|firsthop|> & <|secondhop|>  \\
\multicolumn{1}{c}{}                    & \textbf{GrailQA} & 1194         & 308    & 133                  \\
\multicolumn{1}{c}{}                    & \textbf{WebQSP}  & 4309         & 103    & 1546                 \\ \hline
\multirow{3}{*}{\textbf{DB}}                     &         &              & <|tb\_cl|>     & <|tb\_cl\_vl|> \\
                                        & \textbf{Spider}  & 2726         & 38        & 22           \\
                                        & \textbf{WikiSQL} & 1516         & 18                & 21    \\                      
\bottomrule
\end{tabular}
}
\caption{The average numbers of candidate logical form and two types of primitives in each dataset. LF represents logical form}
\label{number_candidate}
\end{table}

\textbf{Candidate Size}
To better understand the benefit of enumerating primitives in reducing the size of candidates, we compare the numbers of enumerated primitives and logical forms on both KB and DB. In Table \ref{number_candidate}, we show that in KBQA datasets, the number of our primitives is three times less than the number of logical forms used in previous SOTAs~\cite{ye2021rng,gu2021beyond}. In DB where the number of operations is usually larger, this efficiency advantage is more obvious. While a logical form in KB is usually composed of no more than two entities and two hops, a logical form in DB is generally more complex with some uncertainty and the number of primitives and operations in a logical form would be larger. Specifically, many frequently used operations can appear in one logical form in SQL, like SELECT, WHERE, ORDER BY, GROUP BY, JOIN ON, etc and each of them has a unique functionality. In our model, we only need to enumerate two types of primitives (tb\_cl, tb\_cl\_vl) , which are sufficient to generate the complex logical form. As a result, the number of primitives in DB is 30 to 40 times less than that of logical forms. This shows the benefit of enumerating primitives in Uni-Parser is universal across both KB and DB despite their significant differences in structure and complexity. 

\textbf{Case study} For a more intuitive understanding of our model, we show a concrete example to illustrate the results of our model and the logical form enumeration based model RnG-KBQA \cite{ye2021rng} in Figure \ref{figure:case_study}. The top-5 ranked logical forms in RnG-KBQA contain much redundant information, and none of them equals the gold S-expression. In contrast, our output from the ranker is simple and includes the ingredient of the gold s-expression. The output of the ranker is used as the input to the generator. The generator output of the RnG-KBQA is as same as the top-1 logical form from the ranker. This indicates that their generator is more like a correctness rewriter that performs minor edits on the input candidates. In comparison, our model can find the correct first and second hop primitives and generate the correct logical form. It's worth noticing that even though proper primitive is not ranked as the top-1, our generator has the capability to find the correct.

\section{Related Work}
We focus on semantic parsing rather than directly getting the answer, as semantic parsing is more explainable~\cite{zhang2020web,lin2020bridging}. 
Many papers have applied seq2seq models to solve semantic parsing in either KB~\cite{gu2022arcaneqa,ye2021rng} or Table scenarios~\cite{dong2016language,lin2018nl2bash}, treating it as a translation problem that takes a natural question as input and outputs a logical form.

\textbf{Semantic Parsing on KBQA}
Past works have attempted to generate logical forms using a grammar-based bottom-up parser~\cite{berant2013semantic,pasupat2015compositional} or a seq2seq network~\cite{hao2017end,zhang2019complex}. An alternative approach is to produce a list of logical form candidates and then use a ranker to find the ones that best match the intent of the question~\cite{lan2020query,sun2020sparqa,luo2018knowledge}.~\citet{ye2021rng} further employs a generation stage beyond the rank to remedy or supplement existing logical form candidates. Rather than enumerating the complete logical form, our Uni-Parser only gets the relevant primitives, which greatly improves the parser's efficiency and compositional generalization ability. 
Recently, ArcaneQA proposed a generation-based model with dynamic programming induction in the KB search space to improve the faithfulness of the generated programs~\cite{gu2022arcaneqa} but it is still not as accurate as the rank-based model. 
DECAF~\cite{yu2022decaf} jointly generates both logical forms and direct answers, which help them leverage both KB and text to get better final answers. 

\textbf{Semantic Parsing on DBQA (Text2SQL)}
Text2SQL models take both the natural language question and the database table schema as input~\cite{dou2022unisar,zhang2020m}. To get the sequential version of the table schema, prior work commonly linearizes the input as a table name followed by all the column names.~\citet{lin2020bridging,zhang2020f} further show that using table content as supplemental information in Seq2Seq model can provide a better understanding of the table schema. Moreover, it supports the prediction of the conditional part in the logical form as mentioned in~\cite{yavuz2018takes}.~\citet{shaw2020compositional} first shows that the pre-trained Seq2Seq model~\cite{raffel2020exploring} with 3 Billion parameters achieves competitive performance on the Spider dataset.~\cite{scholak2021picard} proposes a constrained decoding method that can be compatible with various large pre-trained language models and achieves promising performance on Spider.  

\textbf{Unified Question Answering}
Many unified QA models convert the structured (KB) or semi-structured data (DB) to unstructured texts, which provides additional information for missing knowledge in the open-domain textual QA by directly linearizing the structured schema into text~\citet{oguz2020unik,xie2022unifiedskg,tay2022unifying}.~\citet{ma2021open} further uses the data-to-text generator to revise the linearized schema into natural language. 
\citet{li2021dual} proposes a hybrid QA model that either answers questions using text or generates the SQL queries from table schema on the textual and Tabular QA datasets. Our Uni-Parser works in a different direction that efficiently parses the questions into the executable logical forms on both KB and DB in a unified framework. 


\section{Conclusion}
For unified semantic parsing on both KB and DB structured data, we propose Uni-Parser, which has three modules: primitive enumeration, ranker, and compositional generator. Our enumeration at the primitive level rather than the logical-form level produces a smaller number of potential candidates, leading to high efficiency in the enumeration and ranker steps. Moreover, training a generator to produce the logical form from the primitives leads to a more generalized and robust compositional generator. Experimental results on both KB and DB QA demonstrate the advantages of Uni-Parser, especially in the compositional and zero-shot settings. 
\section{Acknowledge}
The authors would like to thank the members of Salesforce AI Research team for fruitful discussions, as well as the anonymous reviewers for their helpful feedback.

\section*{Limitations}
The current Uni-Parser model needs to independently train on each of the datasets. In this work, we test it on four datasets. But if having more datasets, this process will be very time costing. A more unified way is having one model trained on all the datasets, either from KB or DB, once and producing a good performance on each dataset.

The other limitation is that the current model needs to indicate whether the question is from KB or DB. This makes the model hard to be applied to reality where which source can answer the question is unknown. Those limitations are challenging and we leave them for further explorations. 

\clearpage

\begin{thebibliography}{53}
\expandafter\ifx\csname natexlab\endcsname\relax\def\natexlab#1{#1}\fi

\bibitem[{Berant et~al.(2013)Berant, Chou, Frostig, and
  Liang}]{berant2013semantic}
Jonathan Berant, Andrew Chou, Roy Frostig, and Percy Liang. 2013.
\newblock Semantic parsing on freebase from question-answer pairs.
\newblock In \emph{Proceedings of the 2013 conference on empirical methods in
  natural language processing}, pages 1533--1544.

\bibitem[{Berant and Liang(2014)}]{berant2014semantic}
Jonathan Berant and Percy Liang. 2014.
\newblock Semantic parsing via paraphrasing.
\newblock In \emph{Proceedings of the 52nd Annual Meeting of the Association
  for Computational Linguistics (Volume 1: Long Papers)}, pages 1415--1425.

\bibitem[{Bogin et~al.(2019)Bogin, Gardner, and Berant}]{bogin2019global}
Ben Bogin, Matt Gardner, and Jonathan Berant. 2019.
\newblock Global reasoning over database structures for text-to-sql parsing.
\newblock \emph{arXiv preprint arXiv:1908.11214}.

\bibitem[{Bollacker et~al.(2008)Bollacker, Evans, Paritosh, Sturge, and
  Taylor}]{bollacker2008freebase}
Kurt Bollacker, Colin Evans, Praveen Paritosh, Tim Sturge, and Jamie Taylor.
  2008.
\newblock Freebase: a collaboratively created graph database for structuring
  human knowledge.
\newblock In \emph{Proceedings of the 2008 ACM SIGMOD international conference
  on Management of data}, pages 1247--1250.

\bibitem[{Chang et~al.(2020)Chang, Yu, Chang, Yang, and Kumar}]{chang2020pre}
Wei-Cheng Chang, Felix~X Yu, Yin-Wen Chang, Yiming Yang, and Sanjiv Kumar.
  2020.
\newblock Pre-training tasks for embedding-based large-scale retrieval.
\newblock \emph{arXiv preprint arXiv:2002.03932}.

\bibitem[{Chen et~al.(2021)Chen, Liu, Yu, Lin, Lou, and
  Jiang}]{chen2021retrack}
Shuang Chen, Qian Liu, Zhiwei Yu, Chin-Yew Lin, Jian-Guang Lou, and Feng Jiang.
  2021.
\newblock Retrack: a flexible and efficient framework for knowledge base
  question answering.
\newblock In \emph{Proceedings of the 59th Annual Meeting of the Association
  for Computational Linguistics and the 11th International Joint Conference on
  Natural Language Processing: System Demonstrations}, pages 325--336.

\bibitem[{Das et~al.(2021)Das, Zaheer, Thai, Godbole, Perez, Lee, Tan,
  Polymenakos, and McCallum}]{das2021case}
Rajarshi Das, Manzil Zaheer, Dung Thai, Ameya Godbole, Ethan Perez, Jay-Yoon
  Lee, Lizhen Tan, Lazaros Polymenakos, and Andrew McCallum. 2021.
\newblock Case-based reasoning for natural language queries over knowledge
  bases.
\newblock \emph{arXiv preprint arXiv:2104.08762}.

\bibitem[{Devlin et~al.(2017)Devlin, Uesato, Bhupatiraju, Singh, Mohamed, and
  Kohli}]{devlin2017robustfill}
Jacob Devlin, Jonathan Uesato, Surya Bhupatiraju, Rishabh Singh, Abdel-rahman
  Mohamed, and Pushmeet Kohli. 2017.
\newblock Robustfill: Neural program learning under noisy i/o.
\newblock In \emph{International conference on machine learning}, pages
  990--998. PMLR.

\bibitem[{Dong and Lapata(2016)}]{dong2016language}
Li~Dong and Mirella Lapata. 2016.
\newblock Language to logical form with neural attention.
\newblock \emph{arXiv preprint arXiv:1601.01280}.

\bibitem[{Dou et~al.(2022)Dou, Gao, Pan, Wang, Lou, Che, and
  Zhan}]{dou2022unisar}
Longxu Dou, Yan Gao, Mingyang Pan, Dingzirui Wang, Jian-Guang Lou, Wanxiang
  Che, and Dechen Zhan. 2022.
\newblock Unisar: A unified structure-aware autoregressive language model for
  text-to-sql.
\newblock \emph{arXiv preprint arXiv:2203.07781}.

\bibitem[{Gu et~al.(2021)Gu, Kase, Vanni, Sadler, Liang, Yan, and
  Su}]{gu2021beyond}
Yu~Gu, Sue Kase, Michelle Vanni, Brian Sadler, Percy Liang, Xifeng Yan, and
  Yu~Su. 2021.
\newblock Beyond iid: three levels of generalization for question answering on
  knowledge bases.
\newblock In \emph{Proceedings of the Web Conference 2021}, pages 3477--3488.

\bibitem[{Gu and Su(2022)}]{gu2022arcaneqa}
Yu~Gu and Yu~Su. 2022.
\newblock Arcaneqa: Dynamic program induction and contextualized encoding for
  knowledge base question answering.
\newblock \emph{COLING}.

\bibitem[{Guo and Gao(2019)}]{guo2019content}
Tong Guo and Huilin Gao. 2019.
\newblock Content enhanced bert-based text-to-sql generation.
\newblock \emph{arXiv preprint arXiv:1910.07179}.

\bibitem[{Hao et~al.(2017)Hao, Zhang, Liu, He, Liu, Wu, and Zhao}]{hao2017end}
Yanchao Hao, Yuanzhe Zhang, Kang Liu, Shizhu He, Zhanyi Liu, Hua Wu, and Jun
  Zhao. 2017.
\newblock An end-to-end model for question answering over knowledge base with
  cross-attention combining global knowledge.
\newblock In \emph{Proceedings of the 55th Annual Meeting of the Association
  for Computational Linguistics (Volume 1: Long Papers)}, pages 221--231.

\bibitem[{He et~al.(2019)He, Mao, Chakrabarti, and Chen}]{he2019x}
Pengcheng He, Yi~Mao, Kaushik Chakrabarti, and Weizhu Chen. 2019.
\newblock X-sql: reinforce schema representation with context.
\newblock \emph{arXiv preprint arXiv:1908.08113}.

\bibitem[{Hwang et~al.(2019)Hwang, Yim, Park, and Seo}]{hwang2019comprehensive}
Wonseok Hwang, Jinyeong Yim, Seunghyun Park, and Minjoon Seo. 2019.
\newblock A comprehensive exploration on wikisql with table-aware word
  contextualization.
\newblock \emph{arXiv preprint arXiv:1902.01069}.

\bibitem[{Lan and Jiang(2020)}]{lan2020query}
Yunshi Lan and Jing Jiang. 2020.
\newblock Query graph generation for answering multi-hop complex questions from
  knowledge bases.
\newblock Association for Computational Linguistics.

\bibitem[{Lan et~al.(2019)Lan, Wang, and Jiang}]{lan2019knowledge}
Yunshi Lan, Shuohang Wang, and Jing Jiang. 2019.
\newblock Knowledge base question answering with topic units.

\bibitem[{Lei et~al.(2022)Lei, Chen, Zhang, Wang, Bansal, Berg, and
  Yu}]{lei2022loopitr}
Jie Lei, Xinlei Chen, Ning Zhang, Mengjiao Wang, Mohit Bansal, Tamara~L Berg,
  and Licheng Yu. 2022.
\newblock Loopitr: Combining dual and cross encoder architectures for
  image-text retrieval.
\newblock \emph{arXiv preprint arXiv:2203.05465}.

\bibitem[{Li et~al.(2021)Li, Ng, Xu, Zhu, Wang, and Xiang}]{li2021dual}
Alexander~Hanbo Li, Patrick Ng, Peng Xu, Henghui Zhu, Zhiguo Wang, and Bing
  Xiang. 2021.
\newblock Dual reader-parser on hybrid textual and tabular evidence for open
  domain question answering.
\newblock \emph{arXiv preprint arXiv:2108.02866}.

\bibitem[{Liang(2013)}]{liang2013lambda}
Percy Liang. 2013.
\newblock Lambda dependency-based compositional semantics.
\newblock \emph{arXiv preprint arXiv:1309.4408}.

\bibitem[{Lin et~al.(2020)Lin, Socher, and Xiong}]{lin2020bridging}
Xi~Victoria Lin, Richard Socher, and Caiming Xiong. 2020.
\newblock Bridging textual and tabular data for cross-domain text-to-sql
  semantic parsing.
\newblock \emph{arXiv preprint arXiv:2012.12627}.

\bibitem[{Lin et~al.(2018)Lin, Wang, Zettlemoyer, and Ernst}]{lin2018nl2bash}
Xi~Victoria Lin, Chenglong Wang, Luke Zettlemoyer, and Michael~D Ernst. 2018.
\newblock Nl2bash: A corpus and semantic parser for natural language interface
  to the linux operating system.
\newblock \emph{arXiv preprint arXiv:1802.08979}.

\bibitem[{Liu et~al.(2021{\natexlab{a}})Liu, Chen, Guo, Lin, and
  Lou}]{liu2021tapex}
Qian Liu, Bei Chen, Jiaqi Guo, Zeqi Lin, and Jian-guang Lou.
  2021{\natexlab{a}}.
\newblock Tapex: Table pre-training via learning a neural sql executor.
\newblock \emph{arXiv preprint arXiv:2107.07653}.

\bibitem[{Liu et~al.(2021{\natexlab{b}})Liu, Hashimoto, Zhou, Yavuz, Xiong, and
  Yu}]{liu2021dense}
Ye~Liu, Kazuma Hashimoto, Yingbo Zhou, Semih Yavuz, Caiming Xiong, and Philip~S
  Yu. 2021{\natexlab{b}}.
\newblock Dense hierarchical retrieval for open-domain question answering.
\newblock \emph{EMNLP}.

\bibitem[{Luo et~al.(2018)Luo, Lin, Luo, and Zhu}]{luo2018knowledge}
Kangqi Luo, Fengli Lin, Xusheng Luo, and Kenny Zhu. 2018.
\newblock Knowledge base question answering via encoding of complex query
  graphs.
\newblock In \emph{Proceedings of the 2018 Conference on Empirical Methods in
  Natural Language Processing}, pages 2185--2194.

\bibitem[{Lyu et~al.(2020)Lyu, Chakrabarti, Hathi, Kundu, Zhang, and
  Chen}]{lyu2020hybrid}
Qin Lyu, Kaushik Chakrabarti, Shobhit Hathi, Souvik Kundu, Jianwen Zhang, and
  Zheng Chen. 2020.
\newblock Hybrid ranking network for text-to-sql.
\newblock \emph{arXiv preprint arXiv:2008.04759}.

\bibitem[{Ma et~al.(2020)Ma, Yan, Pang, Zhang, and Shen}]{ma2020mention}
Jianqiang Ma, Zeyu Yan, Shuai Pang, Yang Zhang, and Jianping Shen. 2020.
\newblock Mention extraction and linking for sql query generation.
\newblock \emph{arXiv preprint arXiv:2012.10074}.

\bibitem[{Ma et~al.(2021)Ma, Cheng, Liu, Nyberg, and Gao}]{ma2021open}
Kaixin Ma, Hao Cheng, Xiaodong Liu, Eric Nyberg, and Jianfeng Gao. 2021.
\newblock Open domain question answering over virtual documents: A unified
  approach for data and text.
\newblock \emph{arXiv preprint arXiv:2110.08417}.

\bibitem[{Oguz et~al.(2020)Oguz, Chen, Karpukhin, Peshterliev, Okhonko,
  Schlichtkrull, Gupta, Mehdad, and Yih}]{oguz2020unik}
Barlas Oguz, Xilun Chen, Vladimir Karpukhin, Stan Peshterliev, Dmytro Okhonko,
  Michael Schlichtkrull, Sonal Gupta, Yashar Mehdad, and Scott Yih. 2020.
\newblock Unik-qa: Unified representations of structured and unstructured
  knowledge for open-domain question answering.
\newblock \emph{arXiv preprint arXiv:2012.14610}.

\bibitem[{Pasupat and Liang(2015)}]{pasupat2015compositional}
Panupong Pasupat and Percy Liang. 2015.
\newblock Compositional semantic parsing on semi-structured tables.
\newblock \emph{arXiv preprint arXiv:1508.00305}.

\bibitem[{Raffel et~al.(2020)Raffel, Shazeer, Roberts, Lee, Narang, Matena,
  Zhou, Li, Liu et~al.}]{raffel2020exploring}
Colin Raffel, Noam Shazeer, Adam Roberts, Katherine Lee, Sharan Narang, Michael
  Matena, Yanqi Zhou, Wei Li, Peter~J Liu, et~al. 2020.
\newblock Exploring the limits of transfer learning with a unified text-to-text
  transformer.
\newblock \emph{J. Mach. Learn. Res.}, 21(140):1--67.

\bibitem[{Scholak et~al.(2021)Scholak, Schucher, and
  Bahdanau}]{scholak2021picard}
Torsten Scholak, Nathan Schucher, and Dzmitry Bahdanau. 2021.
\newblock Picard: Parsing incrementally for constrained auto-regressive
  decoding from language models.
\newblock \emph{arXiv preprint arXiv:2109.05093}.

\bibitem[{Shaw et~al.(2020)Shaw, Chang, Pasupat, and
  Toutanova}]{shaw2020compositional}
Peter Shaw, Ming-Wei Chang, Panupong Pasupat, and Kristina Toutanova. 2020.
\newblock Compositional generalization and natural language variation: Can a
  semantic parsing approach handle both?
\newblock \emph{arXiv preprint arXiv:2010.12725}.

\bibitem[{Sun et~al.(2020)Sun, Zhang, Cheng, and Qu}]{sun2020sparqa}
Yawei Sun, Lingling Zhang, Gong Cheng, and Yuzhong Qu. 2020.
\newblock Sparqa: skeleton-based semantic parsing for complex questions over
  knowledge bases.
\newblock In \emph{Proceedings of the AAAI Conference on Artificial
  Intelligence}, volume~34, pages 8952--8959.

\bibitem[{Tay et~al.(2022)Tay, Dehghani, Tran, Garcia, Bahri, Schuster, Zheng,
  Houlsby, and Metzler}]{tay2022unifying}
Yi~Tay, Mostafa Dehghani, Vinh~Q Tran, Xavier Garcia, Dara Bahri, Tal Schuster,
  Huaixiu~Steven Zheng, Neil Houlsby, and Donald Metzler. 2022.
\newblock Unifying language learning paradigms.
\newblock \emph{arXiv preprint arXiv:2205.05131}.

\bibitem[{Thakur et~al.(2020)Thakur, Reimers, Daxenberger, and
  Gurevych}]{thakur2020augmented}
Nandan Thakur, Nils Reimers, Johannes Daxenberger, and Iryna Gurevych. 2020.
\newblock Augmented sbert: Data augmentation method for improving bi-encoders
  for pairwise sentence scoring tasks.
\newblock \emph{arXiv preprint arXiv:2010.08240}.

\bibitem[{Wang et~al.(2019)Wang, Shin, Liu, Polozov, and
  Richardson}]{wang2019rat}
Bailin Wang, Richard Shin, Xiaodong Liu, Oleksandr Polozov, and Matthew
  Richardson. 2019.
\newblock Rat-sql: Relation-aware schema encoding and linking for text-to-sql
  parsers.
\newblock \emph{arXiv preprint arXiv:1911.04942}.

\bibitem[{Xie et~al.(2022)Xie, Wu, Shi, Zhong, Scholak, Yasunaga, Wu, Zhong,
  Yin, Wang et~al.}]{xie2022unifiedskg}
Tianbao Xie, Chen~Henry Wu, Peng Shi, Ruiqi Zhong, Torsten Scholak, Michihiro
  Yasunaga, Chien-Sheng Wu, Ming Zhong, Pengcheng Yin, Sida~I Wang, et~al.
  2022.
\newblock Unifiedskg: Unifying and multi-tasking structured knowledge grounding
  with text-to-text language models.
\newblock \emph{arXiv preprint arXiv:2201.05966}.

\bibitem[{Xiong et~al.(2020)Xiong, Xiong, Li, Tang, Liu, Bennett, Ahmed, and
  Overwijk}]{xiong2020approximate}
Lee Xiong, Chenyan Xiong, Ye~Li, Kwok-Fung Tang, Jialin Liu, Paul Bennett,
  Junaid Ahmed, and Arnold Overwijk. 2020.
\newblock Approximate nearest neighbor negative contrastive learning for dense
  text retrieval.
\newblock \emph{arXiv preprint arXiv:2007.00808}.

\bibitem[{Yavuz et~al.(2018)Yavuz, G{\"u}r, Su, and Yan}]{yavuz2018takes}
Semih Yavuz, Izzeddin G{\"u}r, Yu~Su, and Xifeng Yan. 2018.
\newblock What it takes to achieve 100\% condition accuracy on wikisql.
\newblock In \emph{Proceedings of the 2018 Conference on Empirical Methods in
  Natural Language Processing}, pages 1702--1711.

\bibitem[{Ye et~al.(2020)Ye, Chen, Wang, Dillig, and Durrett}]{ye2020sketch}
Xi~Ye, Qiaochu Chen, Xinyu Wang, Isil Dillig, and Greg Durrett. 2020.
\newblock Sketch-driven regular expression generation from natural language and
  examples.
\newblock \emph{Transactions of the Association for Computational Linguistics},
  8:679--694.

\bibitem[{Ye et~al.(2021)Ye, Yavuz, Hashimoto, Zhou, and Xiong}]{ye2021rng}
Xi~Ye, Semih Yavuz, Kazuma Hashimoto, Yingbo Zhou, and Caiming Xiong. 2021.
\newblock Rng-kbqa: Generation augmented iterative ranking for knowledge base
  question answering.
\newblock \emph{arXiv preprint arXiv:2109.08678}.

\bibitem[{Yih et~al.(2015)Yih, Chang, He, and Gao}]{yih2015semantic}
Scott Wen-tau Yih, Ming-Wei Chang, Xiaodong He, and Jianfeng Gao. 2015.
\newblock Semantic parsing via staged query graph generation: Question
  answering with knowledge base.
\newblock In \emph{Proceedings of the Joint Conference of the 53rd Annual
  Meeting of the ACL and the 7th International Joint Conference on Natural
  Language Processing of the AFNLP}.

\bibitem[{Yih et~al.(2016)Yih, Richardson, Meek, Chang, and Suh}]{yih2016value}
Wen-tau Yih, Matthew Richardson, Christopher Meek, Ming-Wei Chang, and Jina
  Suh. 2016.
\newblock The value of semantic parse labeling for knowledge base question
  answering.
\newblock In \emph{Proceedings of the 54th Annual Meeting of the Association
  for Computational Linguistics (Volume 2: Short Papers)}, pages 201--206.

\bibitem[{Yu et~al.(2022)Yu, Zhang, Ng, Zhu, Li, Wang, Hu, Wang, Wang, and
  Xiang}]{yu2022decaf}
Donghan Yu, Sheng Zhang, Patrick Ng, Henghui Zhu, Alexander~Hanbo Li, Jun Wang,
  Yiqun Hu, William Wang, Zhiguo Wang, and Bing Xiang. 2022.
\newblock Decaf: Joint decoding of answers and logical forms for question
  answering over knowledge bases.
\newblock \emph{arXiv preprint arXiv:2210.00063}.

\bibitem[{Yu et~al.(2018)Yu, Zhang, Yang, Yasunaga, Wang, Li, Ma, Li, Yao,
  Roman et~al.}]{yu2018spider}
Tao Yu, Rui Zhang, Kai Yang, Michihiro Yasunaga, Dongxu Wang, Zifan Li, James
  Ma, Irene Li, Qingning Yao, Shanelle Roman, et~al. 2018.
\newblock Spider: A large-scale human-labeled dataset for complex and
  cross-domain semantic parsing and text-to-sql task.
\newblock \emph{arXiv preprint arXiv:1809.08887}.

\bibitem[{Zhang et~al.(2019{\natexlab{a}})Zhang, Cai, Xu, and
  Wang}]{zhang2019complex}
Haoyu Zhang, Jingjing Cai, Jianjun Xu, and Ji~Wang. 2019{\natexlab{a}}.
\newblock Complex question decomposition for semantic parsing.
\newblock In \emph{Proceedings of the 57th Annual Meeting of the Association
  for Computational Linguistics}, pages 4477--4486.

\bibitem[{Zhang et~al.(2019{\natexlab{b}})Zhang, Yu, Er, Shim, Xue, Lin, Shi,
  Xiong, Socher, and Radev}]{zhang2019editing}
Rui Zhang, Tao Yu, He~Yang Er, Sungrok Shim, Eric Xue, Xi~Victoria Lin, Tianze
  Shi, Caiming Xiong, Richard Socher, and Dragomir Radev. 2019{\natexlab{b}}.
\newblock Editing-based sql query generation for cross-domain context-dependent
  questions.
\newblock \emph{arXiv preprint arXiv:1909.00786}.

\bibitem[{Zhang and Balog(2020)}]{zhang2020web}
Shuo Zhang and Krisztian Balog. 2020.
\newblock Web table extraction, retrieval, and augmentation: A survey.
\newblock \emph{ACM Transactions on Intelligent Systems and Technology (TIST)},
  11(2):1--35.

\bibitem[{Zhang et~al.(2020{\natexlab{a}})Zhang, Yin, Ma, Ge, and
  Xiao}]{zhang2020f}
Xiaoyu Zhang, Fengjing Yin, Guojie Ma, Bin Ge, and Weidong Xiao.
  2020{\natexlab{a}}.
\newblock F-sql: fuse table schema and table content for single-table text2sql
  generation.
\newblock \emph{IEEE Access}, 8:136409--136420.

\bibitem[{Zhang et~al.(2020{\natexlab{b}})Zhang, Yin, Ma, Ge, and
  Xiao}]{zhang2020m}
Xiaoyu Zhang, Fengjing Yin, Guojie Ma, Bin Ge, and Weidong Xiao.
  2020{\natexlab{b}}.
\newblock M-sql: Multi-task representation learning for single-table text2sql
  generation.
\newblock \emph{IEEE Access}, 8:43156--43167.

\bibitem[{Zhong et~al.(2017)Zhong, Xiong, and Socher}]{zhong2017seq2sql}
Victor Zhong, Caiming Xiong, and Richard Socher. 2017.
\newblock Seq2sql: Generating structured queries from natural language using
  reinforcement learning.
\newblock \emph{arXiv preprint arXiv:1709.00103}.

\end{thebibliography}




\end{document}